\documentclass{article}
\usepackage{spconf,amsmath,graphicx}
\usepackage{xcolor}
\usepackage{amsfonts}

\usepackage{comment}
\usepackage{indentfirst}

\title{Independent Sign Language Recognition \\ with 3D Body, Hands, and Face Reconstruction}
\name{
Agelos Kratimenos\textsuperscript{1,3}, 
Georgios Pavlakos\textsuperscript{2}
and Petros Maragos\textsuperscript{1,3}}

\address{
\textsuperscript{1} School of ECE National Technical University of Athens, 15773 Athens, Greece \\         
\textsuperscript{2} School of EECS University of California Berkeley, California, USA \\
\textsuperscript{3} Robot Perception and Interaction Unit, Athena Research Center, 15125 Maroussi, Greece \\
\small{ageloskrat@yahoo.gr, pavlakos@berkeley.edu, maragos@cs.ntua.gr}
}

\begin{document}
\ninept

\maketitle

\begin{abstract}
Independent Sign Language Recognition is a complex visual recognition problem that combines several challenging tasks of Computer Vision due to the necessity to exploit and fuse information from hand gestures, body features and facial expressions. While many state-of-the-art works have managed to deeply elaborate on these features independently, to the best of our knowledge, no work has adequately combined all three information channels to efficiently recognize Sign Language. In this work, we employ SMPL-X, a contemporary parametric model that enables joint extraction of 3D body shape, face and hands information from a single image. We use this holistic 3D reconstruction for SLR, demonstrating that it leads to higher accuracy than recognition from raw RGB images and their optical flow fed into the state-of-the-art I3D-type network for 3D action recognition and from 2D Openpose skeletons fed into a Recurrent Neural Network. Finally, a set of experiments on the body, face and hand features showed that neglecting any of these, significantly reduces the classification accuracy, proving the importance of jointly modeling body shape, facial expression and hand pose for Sign Language Recognition.
\end{abstract}


\section{Introduction}
\label{sec:intro}

With the term "Sign Language" we refer to a language that employs signs made with the hands and other movements, including facial expressions and postures of the body, used primarily by people who are deaf. Humans are able, due to their natural ability, to identify continuous and independent sign language after they have been trained to identify and understand it. Unfortunately, that is not the case with computers since Sign Language Recognition (SLR) is considered a very hard task due to the need of combining information from three different channels; face, body and hands.  \par
Each independent task has already been successful in the past. But combining all these features in the requisite detail for SLR is far from being perfected. Using raw frames as input has been rigorously tested since it includes all necessary information. However, it is extremely hard for the network to focus on the proper features, especially when the environment alters significantly, while it is also extremely time-consuming and resource-expensive. \par
The importance of SLR is indisputable and managing to train a tool that can efficiently and sufficiently identify Sign Language can be extremely beneficial for the Sign Language community. To eliminate the constrains that raw frames include, we investigate 3D body reconstruction, aiming to
encompass information coming from gestures, facial expressions and body structure.
We experiment with SMPL-X, a sophisticated parametric body model that can reconstruct with notable precision the human body from a single RGB image. In this paper, we compare i) raw images ii) 2D skeleton reconstruction with Openpose and iii) 3D body reconstruction with SMPL-X, to evaluate the efficiency of our proposed method. A secondary contribution of this work is the study of the connection between the three channels of information (i.e., hand pose, facial expressions, body pose) and their importance in SLR. Many works have already highlighted the significance of adding facial expressions and body structure features to the gesture features, 
since many similar words in Sign Language have exactly the same gesture sequence differing only 
in the expression of the face or in the positioning of the hands with regard to the body. We conduct experiments to 
show that the optimal results are achieved only after all three channels of information are being included. \par
The rest of this paper is organized as follows: Section 2 provides a review of related work in the recent literature. The proposed SMPL-X method and its parameters that are being used for SLR, are analyzed in Section 3. Section 4 deals with the experimental setup, describing the sign language dataset used, the feature extraction, their pre-processing and the model architectures we exploit. In Section 5 we present the results of our experiments and evaluate the three techniques and finally, in Section 6 we conclude our remarks and propose plausible future directions.

\begin{figure*}[!htb]
\minipage{0.245\textwidth}
  \includegraphics[width=\linewidth,height=140pt]{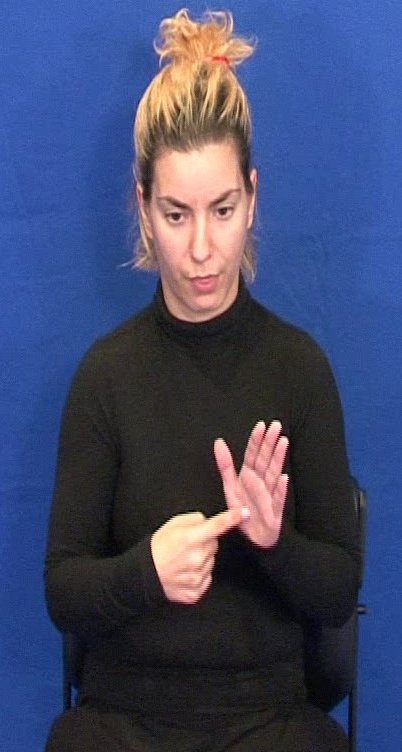}
\endminipage\hfill
\minipage{0.245\textwidth}
  \includegraphics[width=\linewidth,height=140pt]{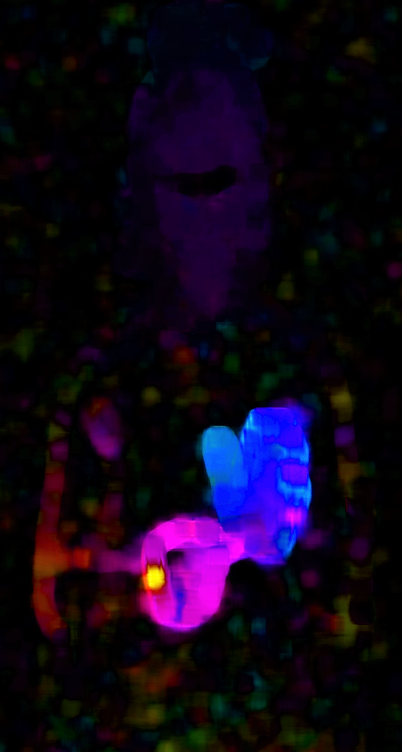}
\endminipage\hfill
\minipage{0.245\textwidth}%
  \includegraphics[width=\linewidth,height=140pt]{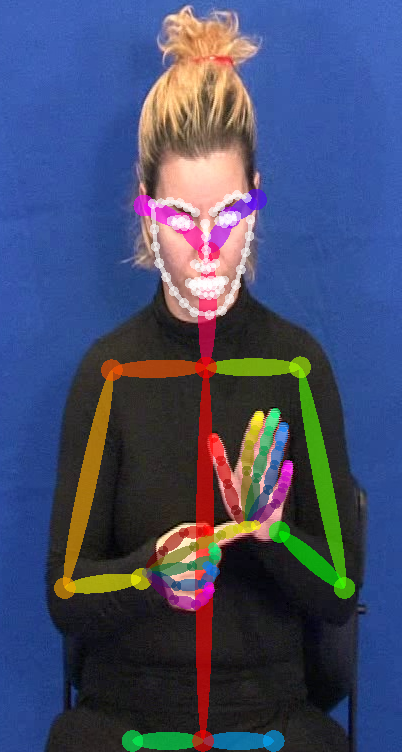}
\endminipage\hfill
\minipage{0.245\textwidth}%
  \includegraphics[width=\linewidth,height=140pt]{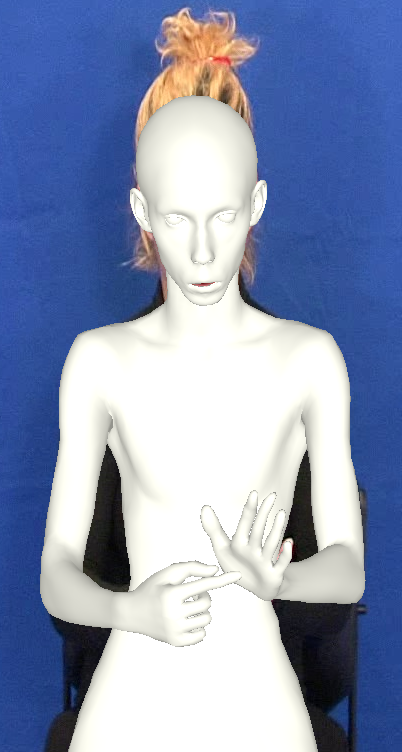}
\endminipage\hfill
\caption{i) First image: Raw RGB frame, ii) Second Image: Optical flow of a frame, iii) Third Image: Openpose 2D Skeleton, iv) Fourth image: 3D Body Reconstruction produced by SMPL-X.}
\vspace{-0.2cm}
\end{figure*}

\section{Related Work}
\label{sec:related}

Sign languages are
natural languages communicable purely by vision  via sequences of time-varying 3D shapes. They  serve for communication in the Deaf communities, as well as among deaf and hearing people if the latter learn to sign. 
They convey information and meaning via spatio-temporal visual patterns, which are formed by manual (handshapes)  and non-manual cues (facial expressions and upper body motion).
Computer-based processing and recognition of sign videos is also broadly related to vision-based human-computer and human-robot interaction  using gesture recognition. \par
While significant progress exists in the field of automatic sign language recognition
 from the computer vision and pattern recognition fields,
(e.g., see \cite{VoMe01, OnRa05, Agr+08, TPM14, KZNB18} and the references therein),
 it still remains a quite
challenging task especially for continuous sign language. In addition to signs
having a complex multi-cue 4D space-time structure, the difficulty in their
automatic recognition is also due to the large variability
with respect to inter-signer or intra-signer variations of signing while expressing the same concept-word.
Due to the above variability, instead of recognizing each sign as a whole `visual word', a
more efficient approach (inspired by speech recognition) is to decompose signs into \emph{subunits},
resembling the phonemes of speech, and recognize them as a specific sequence of subunits by
using some statistical model, e.g., via Hidden Markov Models (HMMs).
Clearly, the subunits approach performs much better on large vocabularies and continuous language;
further, the subunits are  reusable and help with signer adaptation.
In lack of a lexicon, computational techniques that find such subunits are \emph{data-driven},
i.e., perform unsupervised clustering on a large database and use the cluster centroids as subunits.
This performs well  in several instances, especially when the subunits are pre-classified and
statistically modeled based on visual features into dynamic vs. static, as done in~\cite{TPM14}.
A further improved  performance accompanied with phonetic interpretability may be  obtained if the chosen subunits
are also based on the phonetic structure of a sign,
as for example by incorporating
the Posture-Detention-Transition-Steady Shift (PDTS)
system \cite{JoLi11} of phonetic labels.
In \cite{PTVM11}  the phonetic information provided by the PDTS transcriptions of sign videos was combined with
the automatically extracted visual features to create statistically trained
\emph{phonetic subunits} and a corresponding lexicon, which were then used for optimally aligning (via Viterbi decoding) the data with the phonetic labels
and hence providing the missing temporal segmentation, as well as better sign recognition. \par
While information and meaning in sign languages are mainly conveyed by moving handshapes, they are also 
communicated
in part by non-manual cues such as facial expressions. 
These expressions can be visually modeled by deformable  models  that encode both geometric shape and brightness texture information.
Deformable masks provided by active appearance models (AAMs) \cite{CET01} can successfully help with detecting and tracking several types of informative events 
of a sign sequence, e.g., eye blinking, as done in \cite{APM14}. AAMs \cite{RTPM13} have also significantly boosted the performance of handshape recognition in sign language videos.
\par
With the advancement of deep neural networks, much progress has been made in independent and continuous SLR, with the use of CNN and LSTM networks. Koller et al.~\cite{cvpr17} used a pretrained GoogleNet CNN architecture followed by two Bidirectional LSTM layers to achieve, a minimum of 26.8\% word error rate in the  RWTH-PHOENIX-Weather 2014 continuous sign language dataset \cite{phoenix}. In \cite{ms-asl}, Joze and Koller have experimented with different deep learning methods in independent SLR, like the I3D \cite{i3d} that consists of a plethora of Inception Conv3D modules, or the hierarchical co-occurrence network (HCN) \cite{hcn} for body key-points. Moreover, Koller et al.~\cite{pami19} combined CNN-LSTM architectures with multi-stream HMMs in the context of weakly supervised learning for sing language videos.
More recently, Camgoz et al.~\cite{cvpr2020} utilized transformers for the task of SLR, while in \cite{eccvw20} they proposed a multi-channel transformer architecture that combined information for body, face and hands.  

\begin{figure*}[!htb]
\minipage{0.49\textwidth}
  \includegraphics[width=\linewidth]{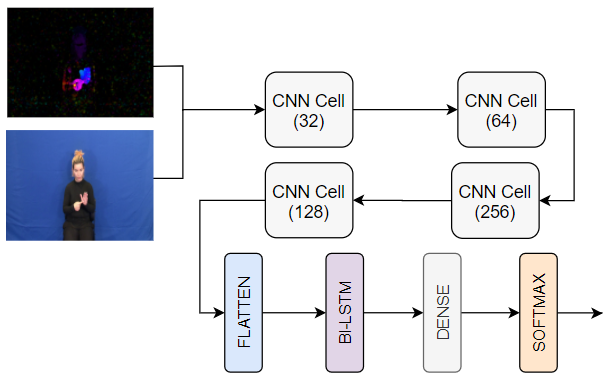}
\endminipage\hfill
\minipage{0.49\textwidth}%
  \includegraphics[width=\linewidth]{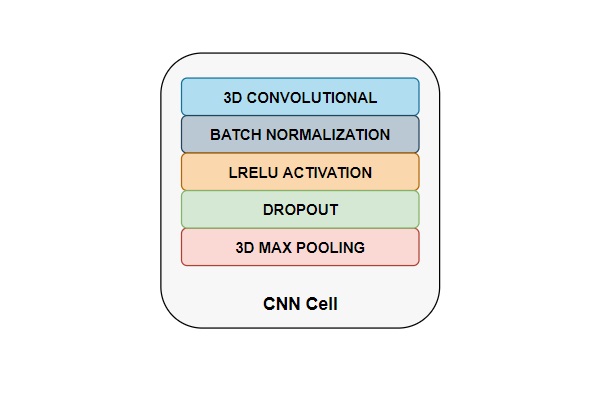}
\endminipage\hfill
\caption{The architecture used for the Convolutional I3D-type model. On the left is the proposed arcitecture with the 3D CNN Cells followed by one Bidirectional LSTM layer. On the right are the interior layers of each 3D CNN cell.}
\end{figure*}

\par
An integral component of our approach is the use
of a recently introduced parametric body model, 
SMPL-X~\cite{pavlakos2019expressive}
that can jointly model the body, 
the hands and the face of the person.
With the exception of Adam~\cite{joo2018total},
this is the only available model
that can jointly capture these
three channels of information.
Previous statistical models, focus only on
the body (e.g., SCAPE~\cite{anguelov2005scape} 
and SMPL~\cite{loper2015smpl}), or add hands,
but still miss the facial expression
(e.g., SMPL+H~\cite{romero2017embodied}), 
which is crucial for the task of sign language
recognition.
Conveniently, SMPL-X is also accompanied by
a method that allows us to reconstruct the model
parameters for a person from a single image.
The method is called SMPLify-X and is based
on the SMPLify approach by Bogo et al. \cite{bogo2016keep}.

\section{Technical Approach}
\label{sec:technical}

In this Section, we provide a short introduction to the SMPL-X~\cite{pavlakos2019expressive} model and then give some details about the SMPLify-X method that enables the estimation of SMPL-X model parameters given a single RGB image as input. Next, we present the dataset we employ to evaluate our approach and finally we analyze the feature extraction methods, the different model architectures and the training hyper-parameters.

\subsection{SMPL-X model and SMPLify-X reconstruction}\label{sec:smplx}

\textbf{SMPL-X} is based on a template mesh which has $N = 10475$ vertices, and a kinematic tree with a set of 54 joints $J$ which are defined by the kinematic
tree and depend on the body shape parameters $\beta$. The model is controlled by pose parameters $\theta$, shape parameters $\beta$ and expression parameters $\psi$. The {\it pose} parameters $\theta$ correspond to the 3D rotation of each model joint, with respect to its parent in the kinematic tree. With 3 DoF per joint, we require 162 pose parameters. Both {\it shape} parameters and {\it expression} parameters correspond to 10 coefficients of a PCA shape and expression shape respectively that are trained to express the majority of body shapes and facial expressions. Given these parameters as input,
the model defines a function $M$ that returns a mesh
with the selected articulation,
body shape and expression:

\begin{equation}
    M\left( \theta, \beta, \psi \right):\mathbb{R}^{| \beta |,| \theta |,| \psi |}
    \rightarrow 
    \mathbb{R}^{3N}.
\end{equation}

This function $M$ is based on Linear Blend Skinning~\cite{Lewis_lbs} and includes some learnable parameters for the pose, shape and expression spaces. These parameters are trained from alignments of the template mesh to 3D scans, as described in~\cite{pavlakos2019expressive}. \par 
\textbf{SMPLify-X} ~\cite{pavlakos2019expressive} is a two stage method, capable of efficiently and effectively reconstructing humans from image data. In the first stage, a set of features is detected on the image, typically 2D keypoints detected by Openpose ~\cite{cao2019openpose,cao2017realtime,simon2017hand,wei2016convolutional}, which include body keypoints, hand keypoints and facial landmarks. We refer to these detected 2D keypoints with $J_{est}$, and we assume that each keypoint $J_{est,i}$ is detected with confidence $\omega_i$. For the second stage, we fit SMPL-X to these 2D landmarks, encouraging the projection of the 3D joints of the model to agree with the detected 2D locations, which is expressed by the data term:

\begin{equation}
\sum_{joint~i} \omega_i \rho (\Pi_K(R_{\theta}            J(\beta)_i))    - J_{est,i}),
  \label{eq:data_term}
\end{equation}

where $R_{\theta}$ is the rotation applied to the joints based on the pose parameters $\theta$, $\Pi_K$ is the projection operation and $\rho$ is a robust function used for the fitting. The complete objective also includes a penalty for mesh intersections and a set of priors for regularization and is optimized with an annealing scheme for the weights of each term to finally produce the corresponding SMPL-X parameters. \par
Although SMPL-X can reconstruct with very high accuracy the person in a specific sequence, our ultimate goal is to recognize the signs for each image sequence. Our main insight is that the low dimensional parametric representation of SMPL-X should capture the majority of the information that is transmitted during a sign,
i.e., the body pose, the hand pose and the facial expressions. This should make it a very effective intermediate representation for sign language recognition. In specific, SMPL-X takes each frame of a video and reconstructs the 3D body, face and hands of the signer conveying it in 88 parameters, converting each sign to a sequence of vectors of length 88. This sequence of SMPL-X parameters across the frames of a sign is being used as input to an RNN to classify the sign to one of the corresponding categories.

\begin{table}
\begin{center}
\scalebox{0.9}{
\begin{tabular}{|c|c|c|c|c|c|}
\hline
GSLL Subset  & Videos & Frames & TrainSet & DevSet & TestSet \\
\hline\hline
50  classes & 538 & 22808 & 318 & 106 & 114 \\
100 classes & 1038 & 45437 & 618 & 206 & 214\\
200 classes & 2038 & 92599 & 1218 & 406 & 414\\
300 classes & 3038 & 140771 & 1818 & 606 & 614 \\
347 classes & 3464 & 161050 & 2066 & 695 & 703 \\
\hline
\end{tabular}
}
\caption{Statistics for the Greek Sign Language Lemmas Dataset and its respective subsets.}
\vspace{-0.5cm}
\end{center}
\end{table}

\begin{table*}[]
\begin{center}
\begin{tabular}{|c||c|c|c|c|c||c|}
\hline
Method \textbackslash \hspace{0.01cm} GSLL Subset & Subset 50 & Subset 100 & Subset 200 & Subset 300 & Full Dataset & Parameters\\ \hline\hline
3D RGB \& Optical Flow Images   & 90.41\%   & 86.85\%    & 80.79\%    & 71.36\%  &65.95\% & 43.41 million  \\ \hline
2D Openpose Skeleton            & 96.49\%   & 94.39\%    & 93.24\%    & 91.86\%  &88.59\%  & 1.55 million \\ \hline
3D SMPL-X Reconstruction & \textbf{96.52\%} & \textbf{95.87\%} & \textbf{95.41\%} & \textbf{95.28\%} &\textbf{94.77\%}  & 0.88 million \\ \hline
\end{tabular}
\caption{Comparison of the three methods for training: i) Raw RGB images and their Optical Flow ii) Openpose skeleton key-points and iii) 3D Body Reconstruction key-points.}
\vspace{-0.65cm}
\end{center}
\end{table*}

\subsection{Greek Sign Language Lemmas Dataset}

Since our goal is to test the ability of the proposed method to adequately extract 3D hand, face and body features, we limit our approach to non-continuous sign language recognition. Continuous SLR contains syntactic and linguistic structure that is beyond the focus of this work. This means that we exclude from our experimentation datasets like RWTH-PHOENIX-Weather 2014 \cite{phoenix} and SIGNUM \cite{signum} that consist of full sentences. Instead, we focus on the Greek Sign Language Lemmas Dataset  (GSLL) \footnote{To be made public upon publication of the paper.}~\cite{TPM14} which proved to be ideal for our experiments. The MS-ASL \cite{ms-asl} dataset consists of 222 signers and extremely varying backgrounds, which makes it challenging for Conv3D networks to converge. In order to make a more fair comparison between 3D reconstruction and 3D convolutional networks we choose the GSLL dataset which consists of only two signers and 347 different signs (classes) in almost 3500 videos and a steady blue background cloth. Table 1 provides more details for the dataset and our selected subsets.

\subsection{Training Methodology}\label{sec:training}

Independent sign language recognition can be considered a task that is similar to action recognition. Thus we expect similar techniques to work well on SLR too. We decided to not intervene on the length of features' sequence. Thus our features vary in sequence length from a minimum of 10 frames to a maximum of 300. Next, we present the methods with which we confront this problem. \par

\textbf{Openpose}: We extract 411 parameters for each frame and feed the sequence in an RNN consisting of one Bi-LSTM layer of 256 units and a  Dense layer for classifying, after applying standard scaling to our features. We believe that by providing a recurrent network with these features will eliminate any redundant information (e.g background, clothes, lighting) that a raw image contains. \par
\textbf{Raw Image and Optical flow}: A 3D state-of-the-art method for action recognition and signing is the I3D network \cite{ms-asl,i3d}. Table 2 shows the architecture of the 3D convolution module used both for raw RGB frames and for the optical flow of these. We reshape each frame to a $175\times175$ array and normalize its pixels to $[0,1]$. We feed raw images to a VGG16-LSTM model as well which is initialized with Imagenet weights, for further experimentation. \par
\textbf{SMPL-X}: Due to its ability to interpret the structure of the body in detail, we strongly believe that this method will provide key features for this task. 
Moreover, SMPL-X provides 3D information, in comparison to Openpose that results to 2D only keypoints, so the extracted features should be strictly more informative.
This method extracts 88 features per frame, creating a (length of sequence) $\times$ 88 array for each sequence, which is being standard scaled as in the Openpose experiments. Similarly to Openpose, we employ the same neural network architecture
so that we can directly compare the two methods independently of the type of architecture. \par

We train all networks using categorical cross-entropy loss. SGD is used to optimize the loss function, with an initial learning rate of 0.0001 and 10\% decay rate per epoch, while the batch size is set to 1, due to varying sequence length. We perform Learning Rate Reduction and Early Stopping by monitoring the validation loss with a patience of 3 and 5 epochs respectively.

\section{Results and Discussion}

According to Table 3, Openpose and SMPL-X models, which consist of 1.6 and 0.9 million parameters respectively, outperform the Conv3D-LSTM and VGG16-LSTM model, which consist of 43 and 15 million parameters respectively. This can be attributed to the fact that the former two eliminate the redundant information from each frame, keeping only the essential body features. Specifically, VGG16 fails to converge and reduce its loss, achieving an accuracy below 10\% for all classes. This does not come as a surprise to us since Joze and Koller in \cite{ms-asl} have trained a VGG16-LSTM model for the MS-ASL dataset which achieved 13.33\% for the ASL100 Subset and just 1.47\% for the ASL500 Subset. As mentioned earlier, GSLL Dataset is characterised by a uniform environment between each sign and each signer (two signers in front of a blue cloth). The MS-ASL dataset consists of 222 distinct signers where each signer performs in a completely altered environment. We strongly believe that Openpose and mainly SMPL-X will by far outperform convolutional models in these datasets, which simulate more accurately the real world. Finally, SMPL-X seems to outperform the features produced by Openpose especially with the increase of different signs, dictating that a more detailed and qualitative representation of the human body is needed for the Sign Language Recognition task. While varying and more complex signs are being added to the train set, Openpose fails to convey the small details that differentiate these signs, while SMPL-X holds its accuracy almost fixed. \par

\par
To further examine the features produced by SMPL-X, we experiment with a combination of a subset of features produced by it. Specifically, as mentioned in Section 3, the SMPL-X method produces a total of 88 features, 10 for shape parameters, 3 for global orientation, 24 for left and right hand pose, 3 for jaw pose, 6 for left and right eye pose, 10 for expression and 32 for the body pose. Moreover, it is widely known that sign language does not depend solely on gestures but fundamentally on body pose and facial expressions as well. To demonstrate this fact, we proceed to a couple of more experiments. First, we remove all information that comes from facial expressions (jaw pose, left and right eye pose and expression) and train the model again with a total of 69 features. Secondly, we only remove the body pose information and train the model with a total of 50 features. Finally, for the sake of completeness, we remove left and right hand pose and train the model with a total of 64 parameters. We conduct the same experiments for Openpose by separating pose keypoints (75 parameters), face keypoints (210 parameters), and left and right hand kepoints (126 parameters). Table 4 sums up the results from all the aforementioned experiments.

\begin{table}
\begin{center}
\begin{tabular}{l|c|c}
\hline
Parameters & Openpose & SMPL-X \\
\hline\hline
All  & \textbf{88.59}\%  & \textbf{94.77}\%\\
Without Face   & 88.34\% & 93.19\%  \\
Without Hands  & 70.20\% & 89.58\%\\
Without Body   & 84.21\% & 85.02\%\\
\end{tabular}
{
  \caption{Experiments with subset of features produced by Openpose and SMPL-X.}
}
\vspace{-0.7cm}
\end{center}
\end{table}

\par First of all, we can see that omitting any of these three channel indeed reduce the accuracy of our model. In fact, we expect the omission of facial characteristics to affect even more the accuracy in the continuous sign language where the face plays a crucial role into expressing the intensity of a word. For example, "rain" and "snow" have the exact same hand configurations, whereas only the mouth shape changes. Furthermore, we observe that removing hand information in SMPL-X is less harmful than removing body pose. That can be attributed to the fact that when few and simple signs are available, the sign can be mainly conveyed through the movement of the arms while the hands often remain straight. Nonetheless, both hands and body structure (chiefly due to arms) are of vital importance for SLR while at the same time, omitting facial expression affect the model's performance. On the other hand, Openpose due to the fact that has very few parameters for body, i.e. only 75 out of 411, it is much more destructive to remove hands features than body.

\section{Conclusion and Future Work}
In this paper we investigated the extraction of 3D body pose, face and hand features for the task of Sign Language Recognition. We compared these features, to Openpose key-points, the most famous method for extracting 2D skeleton parameters and features from raw RGB frames and their optical flow that are fed in a state-of-the-art Deep Learning architecture used in action and sign language recognition. The experiments revealed the superiority of SMPL-X features due to the detailed and qualitative features extraction in the three aforementioned regions of interest. Moreover, we exploited SMPL-X to point out the significance of combining all these three regions for optimal results in SLR. Future work on 3D body, face and hands extraction for SLR includes further experiments in different independent datasets with more signers and varying environment. Furthermore, we strongly believe that applying SMPL-X in continuous SLR will give further prominence to this method, where facial expressions and body structure are even more crucial. Finally, applying SMPL-X in different action recognition tasks is an interesting experiment to examine the universality of SMPL-X success.


\bibliographystyle{IEEEbib}
\bibliography{strings,refs}

\end{document}